\pdfoutput=1

\documentclass[11pt]{article}

\usepackage[review]{acl}

\usepackage{times}
\usepackage{latexsym}

\usepackage[T1]{fontenc}

\usepackage[utf8]{inputenc}

\usepackage{microtype}

\usepackage{inconsolata}

\usepackage{xcolor}
\usepackage{hyperref}
\usepackage{graphicx} 
\usepackage{amsmath}
\usepackage{subfig}
\usepackage{tabularx}
\usepackage{multirow}
\usepackage{tikz}
\usepackage{pgfplots}
\usepackage{bbm}
\usepackage{enumitem}
\usepackage{stfloats}

\pgfplotsset{compat=1.18}

\title{Scalable Detection of Salient Entities in News Articles}

\author{Eliyar Asgarieh \and Kapil Thadani \and Neil O'Hare \\
        Yahoo Research \\
        \texttt{\{eliyar.asgarieh, thadani, nohare\}@yahooinc.com}}

\begin{document}
\maketitle

\begin{abstract}

News articles typically mention numerous entities, a large fraction of which are tangential to the story. Detecting the salience of entities in articles is thus important to applications such as news search, analysis and summarization. In this work, we explore new approaches for efficient and effective salient entity detection by fine-tuning pretrained transformer models with classification heads that use entity tags or contextualized entity representations directly. Experiments show that these straightforward techniques dramatically outperform prior work across datasets with varying sizes and salience definitions. We also study knowledge distillation techniques to effectively reduce the computational cost of these models without affecting their accuracy. Finally, we conduct extensive analyses and ablation experiments to characterize the behavior of the proposed models.

\end{abstract}

\section{Introduction}
\begin{figure}[t]
\centering
\tikzstyle{salient} = [rectangle, rounded corners,
                       minimum height=10pt, very thick, draw=red!50!orange, fill=red!50!orange, fill opacity=0.1]
\tikzstyle{tangent} = [rectangle, 
                       minimum height=0pt, inner sep=0pt, very thick,
                       draw=blue!50!cyan]

\begin{tikzpicture}
    \node at (0,9.0) {\includegraphics[width=0.45\textwidth]{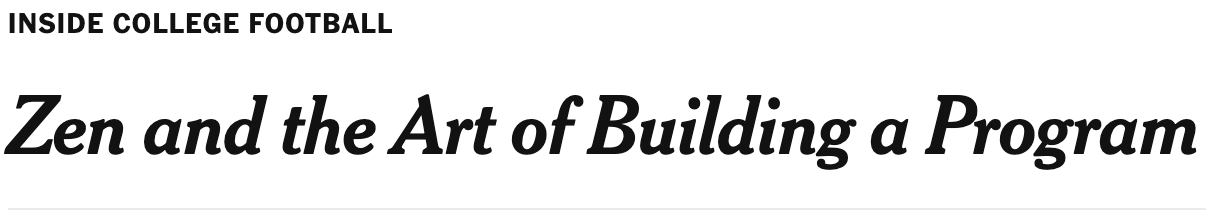}};
    \node at (0,7.8) {\includegraphics[width=0.45\textwidth]{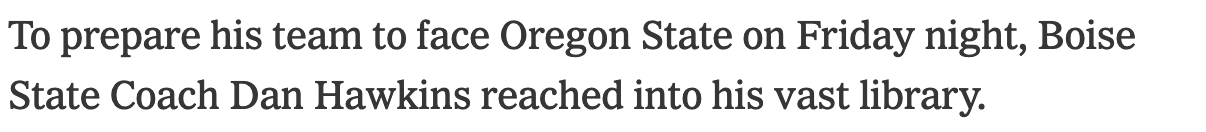}};
    \node at (0,6.0) {\includegraphics[width=0.45\textwidth]{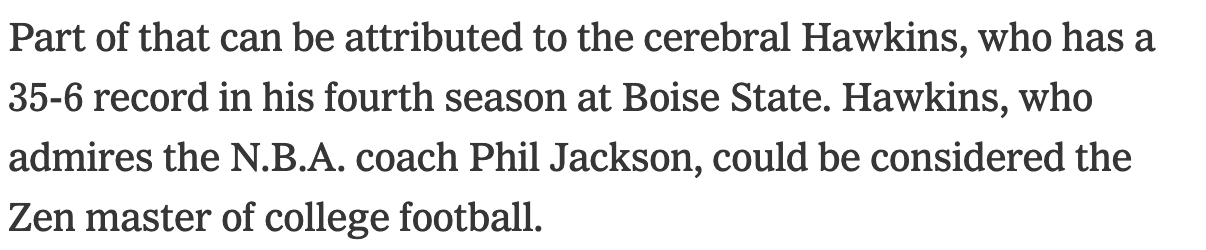}};
    \node at (0,4.0) {\includegraphics[width=0.45\textwidth]{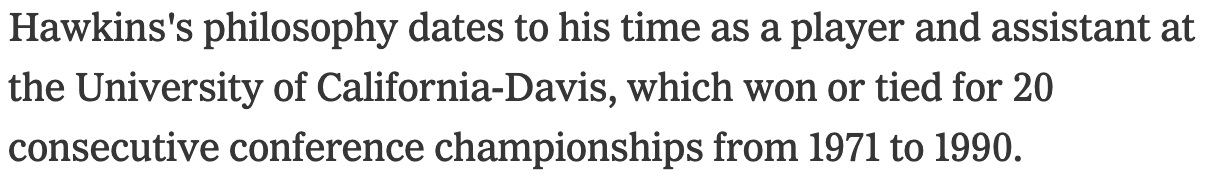}};

    \node at (0,7.2) {$\vdots$};
    \node at (0,5.0) {$\vdots$};

    \node at (-3.2,8.9) [salient, minimum height=15pt, minimum width=25pt] {};
    \node at (-1.5,7.62) [salient, minimum width=43pt] {};
    \node at (1.67,6.53) [salient, minimum width=30pt] {};
    \node at (1.9,6.17) [salient, minimum width=30pt] {};
    \node at (-3.35,5.45) [salient, minimum width=15pt] {};
    \node at (-3.09,4.39) [salient, minimum width=30pt] {};
    \node at (-1.55,4.03) [salient, minimum width=94pt] {};
    
    \node at (0.1,7.8) [tangent, minimum width=42pt] {};
    \node at (2.9,7.8) [tangent, minimum width=18pt] {};
    \node at (-3.27,7.45) [tangent, minimum width=16pt] {};
    \node at (0.7,5.99) [tangent, minimum width=36pt] {};
    \node at (-0.11,5.64) [tangent, minimum width=38pt] {};
    \node at (-1.88,5.64) [tangent, minimum width=20pt] {};

    \draw[black!30, solid] (-3.8,9.9) to (3.8,9.9);
    
    \draw[black!30, solid] (-3.8,9.9) to (-3.8,7.5);
    \draw[black!30, solid] (3.8,9.9) to (3.8,7.5);
    
    \draw[black!30, dotted, thick] (-3.8,7.5) to (-3.8,6.65);
    \draw[black!30, dotted, thick] (3.8,7.5) to (3.8,6.65);

    \draw[black!30, solid] (-3.8,6.65) to (-3.8,5.35);
    \draw[black!30, solid] (3.8,6.65) to (3.8,5.35);

    \draw[black!30, dotted, thick] (-3.8,5.35) to (-3.8,4.5);
    \draw[black!30, dotted, thick] (3.8,5.35) to (3.8,4.5);

    \draw[black!30, solid] (-3.8,4.5) to (-3.8,3.55);
    \draw[black!30, solid] (3.8,4.5) to (3.8,3.55);

    \draw[black!30, dotted, thick] (-3.8,3.55) to (-3.8,3.4);
    \draw[black!30, dotted, thick] (3.8,3.55) to (3.8,3.4);
    

    \node at (-1.5,10.3) {\footnotesize Salient entities};
    \node at (1.5,10.3) {\footnotesize Other entities};
    \node at (-1.5,10.3) [salient, minimum width=58pt, minimum height=12pt] {};
    \node at (1.5,10.1) [tangent, minimum width=50pt] {};
\end{tikzpicture}
\caption{Fragments of a labeled news article from the NYT-Salience dataset. Salient entities (highlighted in orange) include \textit{Dan Hawkins}, \textit{Zen} and \textit{University of California-Davis} while the remaining entities (underlined in blue) are not salient.}
\label{fig:sample_article}
\end{figure}
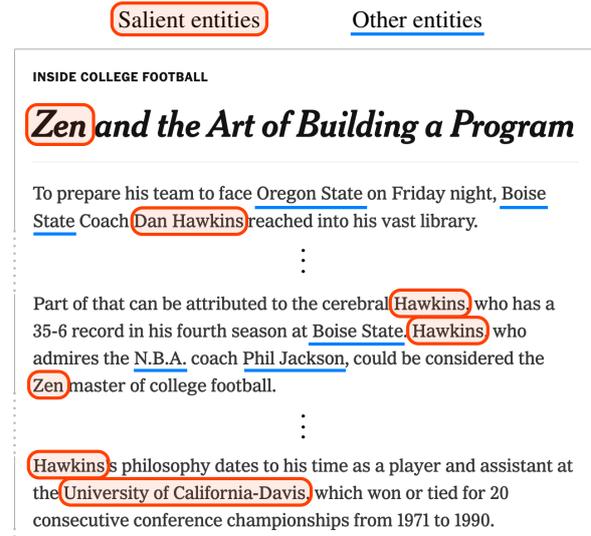

News articles typically mention multiple entities, but are not \textit{about} most of the entities mentioned. This distinction motivates the study of entity \textit{salience} or \textit{aboutness} for news, to identify the entities that are most relevant to the article as would be typically judged by human readers~\cite{gamon2013identifying, dunietz-gillick-2014-new}.
Figure~\ref{fig:sample_article} visualizes salient entity mentions in an annotated New York Times article.
Salient entities can be used to characterize a news story, so they are valuable in systems that index, curate, analyze, or summarize news content.
Consequently, detecting salient entities has been a research focus in multiple domains such as information retrieval \cite{xiong2018towards}, targeting \cite{shi2020salience}, text summarization \cite{sharma-etal-2019-entity,eyal-etal-2019-question,nan-etal-2021-entity}, and relationship extraction \cite{wu2020extracting,sheng2020multi,zhou-etal-2016-attention}.

Research on entity salience estimation is still considered immature, with recent work focused on leveraging better datasets and improving model architectures \cite{dunietz-gillick-2014-new,dojchinovski2016crowdsourced,jain-etal-2020-scirex}. 
The lack of large-scale, high-quality human-annotated data has been one of the main obstacles to developing more advanced and accurate models.
Pretrained transformer encoders such as BERT~\cite{devlin-etal-2019-bert}, however, have enabled the adaptation of capable neural network architectures to tasks for which millions of high-quality examples are not available. Here we examine how, with a simple modification to classification heads, BERT-like models can be adapted to detecting salient entities.

Prior models for this task rely on the computation of meaningful features of entity salience, such as the position of the first mention of an entity, the number of times it appears in article, and the centrality of the entity in a co-occurrence graph. In contrast, the transformer-based models studied here simply require the raw article text and optional
location of candidates, without any feature engineering for the salience task.
 To our knowledge, this is the first application of BERT-like models for estimating entity salience, and we focus on practical concerns for efficient inference. As large transformers can be difficult to run at scale, we use knowledge distillation to compress our salience classifiers to significantly smaller models that require fewer computational resources without diminishing accuracy.
 In addition, because scores from salience classification are often used directly in applications like news recommendation and personalization, we also experiment with improving calibration in distilled models using temperature scaling. 
%
The main contributions of this work are enumerated below: 
\begin{itemize}[itemsep=2pt]
\item We modify classification heads in RoBERTa~\cite{liu2019roberta} variants to model entity salience as a binary classification task. Experiments show that these changes yield state-of-the-art performance on the NYT-Salience~\cite{dunietz-gillick-2014-new}, WN-Salience~\cite{wu2020wnsalience} and SEL-Wikinews~\cite{trani2018sel} datasets.
\item We apply knowledge distillation with teacher ensembles to compress large neural models, training DistillRoBERTa models which are as performant as RoBERTa-Large models that have 4x more parameters.
\item We analyze model performance extensively, showing that the proposed models effectively capture salience signals like position and frequency while generalizing to unseen entities.
\item We examine practical questions such as whether performance transfers across datasets and how temperature scaling during distillation can be used to improve calibration. 
\end{itemize}

\section{Related Work}
Entity extraction has long been an important research topic in information extraction and content understanding, with most work revolving around recognizing entities in text, predicting their relationships, and linking them to knowledge bases. 
In contrast, identifying \textit{which} entities are central to a news story or document remains a more loosely-defined task with unique challenges.

Although there is no single definition of entity salience in the literature, the task is closely related to \textit{keyphrase} extraction in its goal of identifying phrases that are prominent within the context of a document \cite{hasan-ng-2014-automatic,song-etal-2023-survey}.
Keyphrase extraction---along with named entity recognition---can be seen as a candidate generation and pruning stage before salient entity classification; these stages typically entail the use of high-recall techniques to produce candidates for classification \cite{gamon2013identifying, trani2018sel}.

Neural network architectures have a long history in information extraction tasks such as keyphrase extraction, e.g., \citet{sun2021capturing}.
Finding relationships between extracted entities has also been studied extensively \cite{zhou-etal-2016-attention, sheng2020multi, huang-etal-2021-document}, while other work has leveraged deep learning to create comprehensive, end-to-end information extraction models \cite{wu2020extracting, jain-etal-2020-scirex, viswanathan-etal-2021-citationie}. However, we are unaware of prior work that studies modern pretrained neural network architectures such as transformer encoders like BERT~\cite{devlin-etal-2019-bert} to identify salient entities in news.

In the absence of natural labeled data for entity salience, prior work relied on datasets constructed with noisy labels \cite{gamon2013identifying} or heuristics \cite{dunietz-gillick-2014-new, liu-etal-2018-automatic, xiong2018towards} for building and evaluating models.
Datasets with human-annotated salient entities and coreferent mentions have also been published~\cite{dojchinovski2016crowdsourced,trani2018sel,maddela-etal-2022-entsum} but are limited in scale to fewer than a thousand annotated documents each.
Moreover, unsupervised salience estimation techniques~\cite{lu2021salience} perform significantly worse than supervised models.
Creating high-quality datasets for entity salience tasks remains a considerable challenge, given the variation between domains and lack of standardized definitions of the task. 
We benchmark our methods on a variety of datasets to demonstrate robust performance over a wide range of dataset sizes and label definitions.

Designing scalable and practical salient entity detectors remains a challenging information extraction task \cite{jain-etal-2020-scirex}, with prior work based on phrase/document-based hand-engineered features and simple machine learning models \cite{gamon2013identifying, trani2018sel, ponza2019swat}. More recently, kernel-based models using keyphrase embeddings have shown promising results \cite{xiong2018towards, liu-etal-2018-automatic}. 
In applied research, \citet{Appiktala2021}
apply post-processing methods on classifier prediction scores to improve precision.
In this work, we focus on scalable salience classification with powerful pretrained transformer encoders, using modified classification layers and knowledge distillation.

\section{Modeling Salience with Transformers}
Previous work on entity salience~\cite{dunietz-gillick-2014-new, ponza2019swat,wu2020wnsalience} formulates the task as binary classification, using document-based features that need to be computed for each entity candidate $c$ in a document. These manually-defined features usually derive from \citet{dunietz-gillick-2014-new} and incorporate the following signals:
\begin{enumerate}[itemsep=2pt]
    \item \textit{Positional:} The index of the sentence in which entity $c$ appears.
    \item \textit{Count:} Log transformations of the number of occurrences of the head word of the first mention of $c$, and the number of mentions of $c$ including nominal and pronominal references.
    \item \textit{Centrality:} Computed through weighted PageRank over a co-occurrence graph of entities from the training set.
    \item \textit{Lexico-syntactic:} Part-of-speech and lexical features from mentions of $c$.
\end{enumerate}
Notably, transformer architectures~\cite{vaswani2017attention} are capable of capturing all these features---including entity co-occurrence and implicit coreference---from just the text of the article. Text classification benchmarks have been dominated in recent years by pretrained transformer encoders such as BERT~\cite{devlin-etal-2019-bert} and its derivatives. However, formulating salience detection as binary classification for BERT-like models requires the document to be encoded multiple times---once for each candidate entity---which is expensive and inefficient. In this section, we develop two efficient alternatives which enable simultaneous salience estimation for all (or most) entities in a document.

\subsection{Entity classification}
Our proposed approaches modify the use of classification heads in BERT-style transformer encoders. By convention, the first input token for a pretrained encoder is a special classification token \texttt{<CLS>}. During fine-tuning, the encoded representation corresponding to this token from the final transformer layer is used as the input to a multi-layered perceptron (MLP) with a softmax or sigmoid output layer, and the whole network is trained to optimize a classification objective.

Here, we ignore the \texttt{<CLS>} token and instead use an MLP to make classification decisions directly from contextualized representations of each candidate entity from the final transformer layer. Unlike sequence tagging, however, the classifiers have to operate over potentially overlapping multi-token spans.
Although models such as SpanBERT~\cite{joshi-etal-2020-spanbert} are pretrained on a span labeling objective, our methods can be implemented by fine-tuning any BERT-like encoder, which enables distillation to smaller models.
In the following subsections, we examine two different approaches to define fixed-length inputs to classifiers.

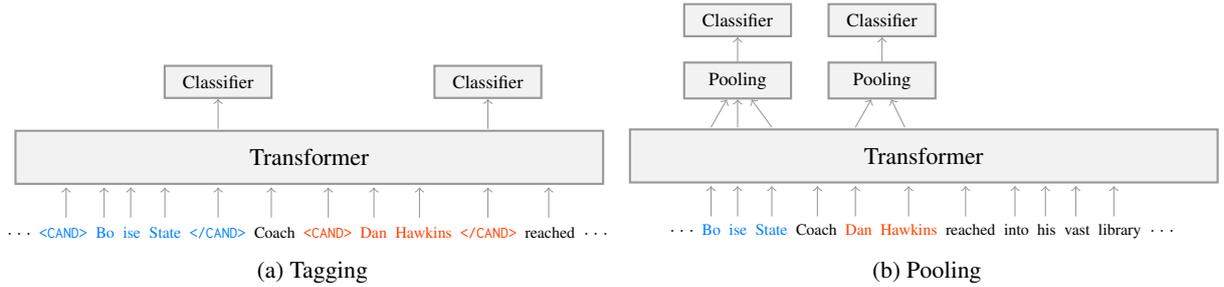
\begin{figure*}[t]
\centering
\tikzstyle{network} = [thick, fill=black!5, draw=black!40]
\tikzstyle{transformer} = [network, minimum width=220pt, minimum height=20pt]
\tikzstyle{vector} = [network]

    \subfloat[Tagging\label{fig:tagging}]{%
        \begin{tikzpicture} 
        \node at (0,2) [transformer] {\footnotesize Transformer};
        \node at (0,1) {\tiny $\cdots$ {\color{blue!50!cyan}\texttt{<CAND>}\; Bo\; ise\; State\; \texttt{</CAND>}\;} Coach\; {\color{red!50!orange}\texttt{<CAND>}\; Dan\; Hawkins\; \texttt{</CAND>}\;} reached\; $\cdots$};

        \node at (-1.2,3) (c1) [vector, minimum width=40pt] {\scriptsize Classifier};
        \node at (2.35,3) (c2) [vector, minimum width=40pt] {\scriptsize Classifier};

        \draw[->, black!40] (-3.2,1.2) to (-3.2,1.6);
        \draw[->, black!40] (-2.7,1.2) to (-2.7,1.6);
        \draw[->, black!40] (-2.35,1.2) to (-2.35,1.6);
        \draw[->, black!40] (-1.9,1.2) to (-1.9,1.6);
        \draw[->, black!40] (-1.2,1.2) to (-1.2,1.6);
        \draw[->, black!40] (-0.5,1.2) to (-0.5,1.6);
        \draw[->, black!40] (0.25,1.2) to (0.25,1.6);
        \draw[->, black!40] (0.85,1.2) to (0.85,1.6);
        \draw[->, black!40] (1.45,1.2) to (1.45,1.6);
        \draw[->, black!40] (2.35,1.2) to (2.35,1.6);
        \draw[->, black!40] (3.15,1.2) to (3.15,1.6);
        
        \draw[->, black!40] (-1.2,2.38) to (c1);
        \draw[->, black!40] (2.35,2.38) to (c2);
        
        \end{tikzpicture} 
    }
    \subfloat[Pooling\label{fig:pooling}]{%
        \begin{tikzpicture} 
        \node at (0,2) [transformer] {\footnotesize Transformer};
        \node at (0,1) {\tiny $\cdots$ {\color{blue!50!cyan}Bo\; ise\; State\;} Coach\; {\color{red!50!orange}Dan\; Hawkins\;} reached\; into\; his\; vast\; library\; $\cdots$};
        
        \node at (-2.45,3) (p1) [vector, minimum width=40pt] {\scriptsize Pooling};
        \node at (-0.55,3) (p2) [vector, minimum width=40pt] {\scriptsize Pooling};
        
        \node at (-2.45,3.8) (c1) [vector, minimum width=40pt] {\scriptsize Classifier};
        \node at (-0.55,3.8) (c2) [vector, minimum width=40pt] {\scriptsize Classifier};
        
        \draw[->, black!40] (-2.8,1.2) to (-2.8,1.6);
        \draw[->, black!40] (-2.45,1.2) to (-2.45,1.6);
        \draw[->, black!40] (-2.0,1.2) to (-2.0,1.6);
        \draw[->, black!40] (-1.4,1.2) to (-1.4,1.6);
        \draw[->, black!40] (-0.9,1.2) to (-0.9,1.6);
        \draw[->, black!40] (-0.2,1.2) to (-0.2,1.6);
        \draw[->, black!40] (0.55,1.2) to (0.55,1.6);
        \draw[->, black!40] (1.20,1.2) to (1.20,1.6);
        \draw[->, black!40] (1.60,1.2) to (1.60,1.6);
        \draw[->, black!40] (2.00,1.2) to (2.00,1.6);
        \draw[->, black!40] (2.50,1.2) to (2.50,1.6);
        
        \draw[->, black!40] (-2.8,2.38) to (p1);
        \draw[->, black!40] (-2.45,2.38) to (p1);
        \draw[->, black!40] (-2.0,2.38) to (p1);
        \draw[->, black!40] (-0.9,2.38) to (p2);
        \draw[->, black!40] (-0.25,2.38) to (p2);

        \draw[->, black!40] (p1) to (c1);
        \draw[->, black!40] (p2) to (c2);
        
        \end{tikzpicture} 
    }
    \caption{Model architectures using tagging and pooling for salience estimation. The classifier layer is a 2-layer MLP with ReLU activations and sigmoid output. Pooling concatenates max- and mean-pooled contextual representations for the highlighted tokens.}
    \label{fig:models}
\end{figure*}

\subsubsection{Tagging entity spans}
In the first approach, which we refer to as \textit{tagging} and illustrate in Figure \ref{fig:tagging}, all candidate entities are tagged with special tokens indicating the start and end of the phrase for which the model is expected to produce a salience prediction. Concretely, let $c_1, \ldots, c_n$ denote the sequence of $n$ subword tokens corresponding to a candidate entity mention $c$. In the tagging approach, distinct start and end tokens \texttt{<CAND>} and \texttt{</CAND>} are inserted before and after each sequence of $c_i$ tokens. The contextualized representation of the \texttt{</CAND>} token for each candidate is used as input for a two-layer MLP which produces binary predictions for salience classification. Formally, the salience score $\hat{y}_c \in [0, 1]$ can be obtained by
\begin{align}
    \hat{y}_c &= \sigma\left(f_\text{tag}\left(r_\text{enc}(c_{n+1})\right)\right)
\end{align}
where $r_\text{enc}(t)$ represents a transformer encoding of some token into its contextualized representation, $f_\text{tag}$ denotes a two-layer MLP, and $\sigma$ denotes a sigmoid activation. The entire network is fine-tuned to minimize the sum of binary cross-entropy losses for each candidate entity in a document.

While this method allows the model to see all candidate entities at prediction time, it has limitations which introduce complexity and cost. First, additional text processing is needed to add special tokens to the input text, and their inclusion reduces the total number of content tokens that can be encoded by the transformer, assuming a fixed maximum sequence length (e.g., 512 tokens). More importantly, the approach cannot process overlapping entities in a single inference pass as there is no natural way to add start and end tags when candidates overlap, e.g., \textit{New York} and \textit{New York Times}. Consequently, the document may need to be re-encoded multiple times to label all entities, adding time and computational cost. 



\subsubsection{Pooling representations}
In the second approach, which we refer to as \textit{pooling} and illustrate in Figure \ref{fig:pooling}, we do not explicitly tag any candidate entities but supply only the article title and body to the encoder. To provide a fixed-length representation of a candidate entity to the classifier, we aggregate contextual representations of all candidate tokens $c_1, \ldots, c_n$ using mean and max pooling, which are then concatenated.
\begin{align}
    r_\text{mean} &= \frac{1}{n} \sum_{i=1}^n r_\text{enc}(c_i) \\
    r_\text{max} &= \max_{1 \leq i \leq n} r_\text{enc}(c_i) \\
    \hat{y}_c &= \sigma\left(f_\text{pool}\left(
    \left[r_\text{mean}, r_\text{max}\right]\right)\right)
\end{align}
where $f_\text{pool}$ is a two-layer MLP for classification.
As in the tagging approach, the entire network is fine-tuned to minimize the sum of binary cross-entropy losses for each candidate entity.

For this approach, no knowledge about the location of candidate entities is needed at the \emph{input} to the model; instead, this is only required at the model output when determining which tokens need to be considered for salience classification. Although the model receives less information at prediction time than in the tagging scenario, the document needs to be encoded just once and overlapping candidates can be classified simultaneously.

As a third, hybrid, approach, we combine pooling and tagging methods by applying pooling on tagged entities. Given the limitations of the tagging approach, this was primarily conducted for the purpose of an ablation study to understand the effect of entity tags on the quality of salience estimation.


\subsection{Knowledge Distillation}
Large transformer models require significant computational resources and add latency, which are obstacles to deploying salience detection at scale. For this reason, we also investigate knowledge distillation techniques to reduce the size of these models with minimal impact to performance metrics~\cite{hinton2015distilling, gou2021knowledge}. Specifically, we use response-based knowledge distillation with teacher ensembles, and we experiment with temperature scaling to transfer knowledge effectively from larger \textit{teacher} models to smaller \textit{student} models.
%

\makeatletter
\newcommand{\hathat}[1]{%
\begingroup%
  \let\macc@kerna\z@%
  \let\macc@kernb\z@%
  \let\macc@nucleus\@empty%
  \hat{\mathchoice%
    {\raisebox{.35ex}{\vphantom{\ensuremath{\displaystyle #1}}}}%
    {\raisebox{.35ex}{\vphantom{\ensuremath{\textstyle #1}}}}%
    {\raisebox{.2ex}{\vphantom{\ensuremath{\scriptstyle #1}}}}%
    {\raisebox{.16ex}{\vphantom{\ensuremath{\scriptscriptstyle #1}}}}%
    \smash{\hat{#1}}}%
\endgroup%
}
\makeatother

Concretely, if $\hathat{y}_c$ denotes the prediction of a student model, these models are trained with a binary cross-entropy loss defined over soft labels $\hat{y}_c \in [0, 1]$ from a teacher model or ensemble, i.e.,
\begin{align}
& - \sum_c \hat{y}_c  \log \hathat{y}_c + \left(1 - \hat{y}_c \right) \log \left (1 - \hathat{y}_c \right)
\end{align}
Both student and teacher predictions contain a temperature term, i.e., $\hat{y} = \sigma(z/T)$ where $T$ is a scalar temperature and the logit $z$ is obtained from the MLP $f_\text{tag}$ or $f_\text{pool}$.
In Appendix~\ref{sec:calibration}, we experiment with the temperature parameters for both teacher and student models in order to control student model calibration \cite{guo2017calibration}. We demonstrate that calibration can be preserved without post-processing using temperature alone, albeit with a small decrease in accuracy.

\begin{table*}[!t]
\renewcommand{\arraystretch}{1.1}
\centering
\small
\begin{tabular}{c|rr|cc|c|cc} 
\hline
{Dataset} & \multicolumn{2}{c|}{{Documents}} & \multicolumn{2}{c|}{{Entities per doc}}  & {Median tokens} & \multicolumn{2}{c}{{\% unique mentions within 512 tokens}} \\
  & \multicolumn{1}{c}{{Train}} & \multicolumn{1}{c|}{{Test}} & {Salient} & {Non-salient} & {per doc} & {Salient} & {Non-salient}\\

\hline
 \ NYT-Salience & 100,834 & 9,706 & 2.6 & 17.4 & 809 & 98.9\% & 89.2\% \\ 
 \ WN-Salience  & 5,928 & 1,040 & 3.0 & 9.4 & 268 & 98.9\% & 97.2\% \\ 
 \ SEL-WikiNews  & \multicolumn{2}{c|}{365} & 9.7 & 18.2 & 260 & 99.9\% & 98.8\% \\ 
 \hline
\end{tabular} 
\caption{Key statistics of the datasets used in this work: NYT-Salience~\cite{dunietz-gillick-2014-new}, WN-Salience~\cite{wu2020wnsalience} and SEL-WikiNews~\cite{trani2018sel}.}
\label{table:corpus_stats}
\end{table*}



\section{Datasets}
We evaluate this work on the following public datasets with salient entity annotations. Table \ref{table:corpus_stats} contains relevant statistics of these datasets.
\begin{enumerate}
\item \textbf{NYT-Salience}~\cite{dunietz-gillick-2014-new} is a large-scale dataset that was constructed automatically by identifying entities in the human-written abstracts from the New York Times corpus~\cite{sandhaus08} and aligning them to entities in the article text. 

\item \textbf{WN-Salience}~\cite{wu2020wnsalience} is sampled from the Wikinews corpus which comes with in-text annotations linking the entities to Wikipedia pages. These annotations are used to identify salient entities, which \citet{wu2020wnsalience} claim provides more accurate salience estimation. In comparison, NYT-Salience documents are longer and contain more entities, but may have noisier labels.

\item \textbf{SEL-Wikinews}~\cite{trani2018sel} is a small dataset drawn from Wikinews that contains crowdsourced human judgments of entity salience on an ordinal scale of 0--3, averaged over multiple annotations. We follow \citet{wu2020wnsalience} and obtain binary salience classes from these scores using a threshold of 2.0.
\end{enumerate}
NYT-Salience and WN-Salience have standard test sets, which make evaluations directly comparable with previous published work, and we use about 5\% of the training splits as a validation set. In contrast, SEL-Wikinews has no fixed test set and was originally evaluated using 5-fold cross-validation~\cite{trani2018sel,ponza2019swat}. For our experiments, we partition the data into 298 training, 29 validation and 38 test documents.

We examined a number of other datasets that were not chosen for our experiments. The Reuters-128 Salience dataset \cite{dojchinovski2016crowdsourced} includes only 128 news articles, which is not adequate to reliably benchmark neural network models. Scirex \cite{jain-etal-2020-scirex} has semi-automated salience annotations but it focuses on scientific text rather than news and also contains only 438 documents. EntSUM~\cite{maddela-etal-2022-entsum} contains manual annotations of entity salience on a subset of the New York Times corpus~\cite{sandhaus08} but has not previously been benchmarked on this task and also contains only 693 documents.

\section{Experiments}
In this section, we present evaluation results on the publicly available benchmark datasets enumerated above.
For comparison wnear-real-time in the salience classification setting, we report precision (P), recall (R) and F$_1$ on each dataset. In addition, we report average precision (AP), which measures the area under the precision-recall curves and provides a single-valued measure of model accuracy that is threshold-independent, unlike F$_1$. An alternative evaluation setting based on ranking metrics is discussed in Appendix~\ref{sec:ranking}.

\subsection{Implementation Details}
We use pretrained RoBERTa models \cite{liu2019roberta} for all experiments, with a maximum input sequence length of 512 tokens. As indicated in Table~\ref{table:corpus_stats}, 512 tokens cover almost all of the salient entities and more than 89\% of the non-salient entities in both datasets. Input text was preprocessed to drop punctuation before tokenization, which we found to improve performance.

We build on the HuggingFace \textit{transformers} library \cite{wolf-etal-2020-transformers} and pretrained models. The classification head is a two-layer MLP with a ReLU activation after the first layer, a hidden layer half the size of the input, and a sigmoid output activation. We use AdamW~\cite{loschilov-hutter-2019-adamw} with weight decay 0.01, epsilon 1e-8 and a typical learning rate of 1e-5.

We apply loss re-weighting to tackle label imbalance in the data---typically biased towards negatives---with a smoothed inverse class-frequency weight similar to \citet{ye2022multilingual}. The loss for each candidate entity $c$ is scaled by
\begin{align}
\label{eq:class_reweighting}
    w_c &= \frac{\max_j f_j + \alpha}{f_i + \alpha}
\end{align}
where $f_i$ denotes the class frequency of label $i \in \{0, 1\}$ in the training data and $\alpha$ is a smoothing term. We calculate $f_i$ per batch of data during training and use $\alpha=0.01$ as the smoothing factor.

\begin{table*}[t]
\centering
\small
\setlength{\tabcolsep}{4.3pt}
\renewcommand{\arraystretch}{1.1}
\begin{tabular}{lr|cccc|cccc|cccc} 
\hline
\multirow{2}{*}{Model} &  & \multicolumn{4}{c|}{NYT-Salience} & \multicolumn{4}{c|}{WN-Salience} & \multicolumn{4}{c}{SEL-Wikinews} \\
& & P & R & F$_1$ & AP & P & R & F$_1$ & AP & P & R & F$_1$ & AP \\
\hline
\hline
 \multicolumn{2}{l|}{\citet{dunietz-gillick-2014-new}} &  0.605 & 0.635 & 0.620  & - & - & - & - & - & - & - & - & - \\
 \multicolumn{2}{l|}{\citet{trani2018sel}} & - & - & - & - & - & - & - & - & 0.50$^\dagger$ & 0.61$^\dagger$ & 0.52$^\dagger$ & - \\ 
 \multicolumn{2}{l|}{\citet{ponza2019swat}}  & 0.624 & 0.660 & 0.641 & - & - & - & - & - & 0.58$^\dagger$ & 0.65$^\dagger$ & 0.61$^\dagger$ & -    \\
\multicolumn{2}{l|}{\citet{wu2020wnsalience}} &  0.560 & 0.410 & 0.473  & - & 0.479 & 0.532 & 0.504 & - & 0.403 & \textbf{0.856} & 0.548 & -   \\
\hline
\multirow{2}{*}{DistilRoBERTa} & Tagging &  0.643 & 0.696 & 0.668 & 0.737  &  0.599 & 0.710 & 0.650 & 0.732 & 0.737 & 0.812 & 0.773 & 0.841  \\
& Pooling &  0.662 & 0.709 & 0.685 & 0.752 & 0.664 & 0.701 & 0.682 & 0.760 & 0.763 & 0.802 & 0.782 & 0.841   \\
\hline
\multirow{2}{*}{RoBERTa-Base} & Tagging & 0.669 & \textbf{0.732} & 0.699 & 0.768 & 0.647 & 0.721 & 0.682 & 0.763 & 0.774 & 0.782 & 0.778 & 0.860   \\
 & Pooling &   0.675 & 0.728 & 0.701 & 0.773 & 0.657 & \textbf{0.725} & 0.689 & 0.772 & 0.797 & 0.838 & 0.817 & 0.859   \\
\hline
\multirow{2}{*}{RoBERTa-Large} & Tagging & \textbf{0.711} & 0.711 & 0.711 & 0.783 & 0.715 & 0.690 & 0.703 & 0.789 & \textbf{0.802} & 0.843 & 0.822 & 0.865  \\
 & Pooling & 0.703 & 0.725 & \textbf{0.714} & \textbf{0.789} & \textbf{0.720} & 0.705 & \textbf{0.712} & \textbf{0.794} & 0.796 & 0.853 & \textbf{0.824} & \textbf{0.876}  \\
\hline
\end{tabular} 
\caption{Performance of the proposed pooling and tagging approaches for RoBERTa models of different sizes, as well as state-of-the-art baselines. $\dagger$ indicates a different evaluation methodology that cannot be compared directly.}
\label{table:model_sizes}
\end{table*}

\begin{table}[t]
\centering
\small
\setlength{\tabcolsep}{1.9pt}
\renewcommand{\arraystretch}{1.1}

\begin{tabular}{l|c|c|c} 
\hline
{Scoring} & NYT-Salience & WN-Salience & SEL-Wikinews \\
\hline
\hline
First mention & \textbf{0.790} & \textbf{0.794} & 0.879 \\
Last mention &  \textbf{0.790} & 0.793 & 0.879 \\
\hline 
Average & \textbf{0.790} & \textbf{0.794} & 0.884 \\
Median &  \textbf{0.790} & \textbf{0.794} & \textbf{0.886} \\
\hline
\end{tabular} 
\caption{Performance (AP) when varying the choice of mentions used to produce salience scores using pooling with RoBERTa-Large.}
\label{table:mention}
\end{table}

\subsection{Comparison against Baselines}
Table \ref{table:model_sizes} compares individual fine-tuned models trained using our proposed classification schemes against prior work. We can see that transformer-based models produce significant improvements over prior techniques based on hand-engineered features. The results unsurprisingly improve as model size increases.
The pooling approach seems to perform best across datasets and metrics, showing that models are able to identify salience just by relying on the content without special tokens. In addition, all candidate entity spans in the document can be scored in a single inference pass in the pooling approach, even if they are overlapping. The best pooling models outperform the strongest baseline models on F$_1$ by over 11\% for NYT-Salience, 41\% for WN-Salience, and 34\% for SEL-Wikinews.

We also compare these approaches against standard RoBERTa classifiers---which re-encode the document for each entity---in Appendix~\ref{sec:standard}. In addition, we discuss the limitations of large language models (LLMs) for our setting in Appendix~\ref{sec:llms}.

\begin{table*}[t]
\centering
\small
\setlength{\tabcolsep}{4.4pt}
\renewcommand{\arraystretch}{1.1}
\begin{tabular}{l|cccc|cccc|cccc} 
\hline
\multirow{2}{*}{Model} & \multicolumn{4}{c|}{NYT-Salience} & \multicolumn{4}{c|}{WN-Salience} & \multicolumn{4}{c}{SEL-Wikinews} \\
& P & R & F$_1$ & AP & P & R & F$_1$ & AP & P & R & F$_1$ & AP \\
\hline
\hline
Tagging & 0.669 & \textbf{0.732} & 0.699 & 0.768 & 0.647 & 0.721 & 0.682 & 0.763 & 0.774 & 0.782 & 0.778 & 0.860   \\
Pooling with tags & 0.667 & 0.728 & 0.696 & 0.768 & 0.666 & 0.677 & 0.672 & 0.754 & \textbf{0.822} & 0.772 & 0.796 & 0.856   \\
Pooling (non-overlapping) & \textbf{0.681} & 0.715 & 0.698 & 0.769 & \textbf{0.697} & 0.694 & \textbf{0.696} &\textbf{0.772} & 0.805 & 0.777 & 0.791 & 0.854  \\
Pooling &   0.675 & 0.728 & \textbf{0.701} & \textbf{0.773} & 0.657 & \textbf{0.725} & 0.689 & \textbf{0.772} & 0.797 & \textbf{0.838} & \textbf{0.817} & \textbf{0.859}    \\
\hline 
\end{tabular} 
\caption{Ablation results for different classification strategies using the RoBERTa-Base model.} 
\label{table:formulations}
\end{table*}


\subsection{Ablation of Model Variants}
For the tagging approach, we need to identify non-overlapping candidates in an article and tag them with special tokens, which will produce a slightly different sample distribution from the models that use pooling and can thus train on overlapping candidates in parallel. To enable better comparisons between these methods, we created an alternative version of the training data for the \emph{Pooling} model that uses the same samples with non-overlapping candidates as the \textit{Tagging} model. In Table \ref{table:formulations}, \textit{Pooling (non-overlapping)} indicates cases in which pooling models are trained on a version of the data with no overlapping entities. In addition, \textit{Pooling with tags} indicates a hybrid approach that uses pooled representations from the tagged version of the data.
The \emph{Pooling (non-overlapping)}, \emph{Pooling with tags}, and \emph{Tagging} approaches all use the same sample distributions, while \emph{Pooling} is the only variant trained with all overlapping candidates in parallel.

All models are evaluated on the same number of entities and documents.
We use the model score of the first mention of an entity for salience estimation. Table~\ref{table:mention} shows that this choice does not affect performance. Notably, aggregating scores from all mentions does not improve over using a single mention, suggesting that the model successfully learns document-level representations of entities.


\begin{table*}[t]
\centering
\small
\setlength{\tabcolsep}{2.8pt}
\renewcommand{\arraystretch}{1.1}
\begin{tabular}{l|r|cccc|cccc|cccc} 
\hline
\multirow{2}{*}{Model} & Num. & \multicolumn{4}{c|}{NYT-Salience} & \multicolumn{4}{c|}{WN-Salience} & \multicolumn{4}{c}{SEL-Wikinews} \\
& params & P & R & F$_1$ & AP & P & R & F$_1$ & AP & P & R & F$_1$ & AP \\
\hline
\hline
DistilRoBERTa & 82M &  0.662 & 0.709 & 0.685 & 0.752 & 0.664 & 0.701 & 0.682 & 0.760 & 0.763 & 0.802 & 0.782 & 0.841   \\
RoBERTa-Base & 125M &   0.675 & 0.728 & 0.701 & 0.773 & 0.657 & 0.725 & 0.689 & 0.772 & 0.797 & 0.838 & 0.817 & 0.859   \\
RoBERTa-Large & 355M & 0.703 & 0.725 & 0.714 & 0.789 & \textbf{0.720} & 0.705 & 0.712 & 0.794 & 0.796 & \textbf{0.853} & 0.824 & 0.876  \\
\hline
$4\times$RoBERTa-Large \textit{(teacher)} & 1.42B & \textbf{0.718} & 0.720 & \textbf{0.719} & \textbf{0.796} & 0.671 & \textbf{0.764} & \textbf{0.715} & 0.792 & 0.827 & 0.827 & \textbf{0.827} & \textbf{0.895}   \\
\hline
DistilRoBERTa \textit{(distilled)} & 82M & 0.702 & 0.706 & 0.704 & 0.775 & 0.685 & 0.715 & 0.699 & 0.784 & \textbf{0.833} & 0.787 & 0.809 & 0.864 \\
RoBERTa-Base \textit{(distilled)} & 125M & 0.703 & \textbf{0.731} & 0.717 & 0.793 & 0.686 & 0.746 & 0.714 & \textbf{0.799} & 0.804 & 0.812 & 0.808 & 0.885 \\
\hline
\end{tabular} 
\caption{Performance of distilled pooling models for NYT-Salience and WN-Salience in comparison to the teacher ensemble ($4\times$RoBERTa-Large) and conventionally-trained models of the same size.}
\label{table:kd_results}
\end{table*}

\subsection{Evaluation of Knowledge Distillation}
While RoBERTa-Large models produce stronger results, their latency and computational needs impede near real-time deployments that have to scale to the daily volume of online news. To maintain accuracy while reducing computational cost, we use knowledge distillation~\cite{hinton2015distilling, gou2021knowledge}. First, we create an ensemble model by averaging the predictions of four RoBERTa-Large models: (1) \textit{Pooling}, (2) \textit{Pooling (non-overlapping)}, (3) \textit{Pooling with tags}, and (4) \textit{Tagging}. Using this ensemble of four models,\footnote{We also trained a more efficient ensemble of four RoBERTa-Large \textit{Pooling} models initialized with different random seeds and obtained similar results to the heterogenous ensemble described here.} each with 355M parameters (1.42B parameters in total), we fine-tune student models based on RoBERTa-Base (125M parameters) and DistilRoBERTa (82M parameters). 

    In this section, we describe distillation experiments for unscaled teacher and student outputs, i.e., student and teacher temperature parameters are set to 1. As shown in Appendix~\ref{sec:transferability}, we did not see much benefit in using teacher model predictions outside the data the model was trained on (e.g., inferring on NYT-Salience using model trained on WN-Salience or vice versa), so we just use in-domain predictions to produce the transfer sets. Although this scenario can also incorporate gold labels in distillation, such as in \citet{wu-etal-2020-scalable}, we do not pursue that approach here in order to simulate a practical scenario of salience detection, which would typically employ a large-scale un-annotated transfer set with more recent and/or in-domain examples.
All distilled models are pooling models so they can be used to infer the salience score of all entities in a document in a single pass. 


The results in Table \ref{table:kd_results} show that the knowledge distillation approach is successful at reducing model size while maintaining accuracy. Focusing on the two larger datasets---as SEL-Wikinews performance is also limited by a small transfer set---we see that
distilled models show improvements over otherwise equivalent models which are fine-tuned conventionally from the base model, with gains of 2.3\% to 3.6\% in F$_1$ and AP.
The distilled RoBERTa-Base models even surpass the
conventional RoBERTa-Large models on F$_1$ and AP for both datasets despite having 65\% fewer parameters. Similarly, the distilled DistilRoBERTa models improve over the conventional RoBERTa-Base models despite having 34\% fewer parameters. These results demonstrate that distillation can be effectively employed to scale up salience detection without performance penalties. In addition, the RoBERTa-Base distillation results in particular indicate that distilling from an ensemble offers benefits over distilling from a single large model when computational resources are available.


\subsection{Stratified Analysis of Results}
In order to better understand the behavior of the proposed models, this section examines model performance when isolated to particular groups of entities. Through this, we can examine whether the models adequately capture well-known signals of salience such as entity position and frequency, and whether they generalize beyond their training data.

\begin{figure}[t]
\begin{tikzpicture}
    \begin{axis}[
        name=ap,
        legend columns=3,
        legend image code/.code={
            \draw[mark repeat=2,mark phase=2]
            plot coordinates {
            (0cm,0cm)
            (0.25cm,0cm)        
            (0.45cm,0cm)         
            };%
        },
        legend style={at={(0.5,1.05)}, anchor=south, draw=none, font=\scriptsize, /tikz/every even column/.append style={column sep=3pt}},
        small,
        width=210pt,
        height=140pt,
        ylabel={AP},
        xmode=log,
        log ticks with fixed point,
        xticklabels={\empty},
        extra x ticks={1, 2, 5, 10, 20, 50, 100},
        extra x tick labels={},
        xmin=1,
        xmax=100,
        ymin=0.65,
        ymax=1,
        xmajorgrids=true,
        ymajorgrids=true,
]

\addplot [blue!50!cyan, thick] coordinates {
    (1, 0.949)
    (2, 0.944)
    (5, 0.924)
    (10, 0.906)
    (20, 0.873)
    (50, 0.820)
    (100, 0.790)
};
\addlegendentry{NYT-Salience}

\addplot [red!80!orange, thick] coordinates {
    (1, 0.933)
    (2, 0.926)
    (5, 0.928)
    (10, 0.911)
    (20, 0.882)
    (50, 0.839)
    (100, 0.794)
};
\addlegendentry{WN-Salience}

\addplot [green!70!black, thick] coordinates {
    (1, 0.967)
    (2, 0.981)
    (5, 0.967)
    (10, 0.946)
    (20, 0.927)
    (50, 0.897)
    (100, 0.879)
};
\addlegendentry{SEL-Wikinews}
\end{axis}

    \begin{axis}[
        name=posrate,
        at={(ap.outer south)}, anchor=outer north,
        legend columns=3,
        legend style={at={(0.5,1.05)}, anchor=south, draw=none, font=\scriptsize, /tikz/every even column/.append style={column sep=3pt}},
        small,
        width=210pt,
        height=120pt,
        xlabel={Offset of first mention (\% of words)},
        ylabel={\% of positive labels},
        xmode=log,
        log ticks with fixed point,
        xticklabels={\empty},
        extra x ticks={1, 2, 5, 10, 20, 50, 100},
        extra x tick labels={\quad 1, \quad 2, 3-5, 6-10, 11-20, 21-50, 51-100},
        extra x tick style={xticklabel style={font=\scriptsize, xshift=0.0ex, anchor=60}},
        xmin=1,
        xmax=100,
        ymin=10,
        ymax=100,
        xmajorgrids=true,
        ymajorgrids=true,
]

\addplot [blue!50!cyan, thick, dashed] coordinates {
    (1, 73.18)
    (2, 72.07)
    (5, 67.10)
    (10, 61.56)
    (20, 53.65)
    (50, 40.73)
    (100, 32.38)
};
\addlegendentry{NYT-Salience}

\addplot [red!80!orange, thick, dashed] coordinates {
    (1, 59.68)
    (2, 60.14)
    (5, 58.34)
    (10, 54.57)
    (20, 48.77)
    (50, 41.14)
    (100, 35.42)
};
\addlegendentry{WN-Salience}

\addplot [green!70!black, thick, dashed] coordinates {
    (1, 95.53)
    (2, 96.20)
    (5, 92.62)
    (10, 86.43)
    (20, 80.26)
    (50, 64.05)
    (100, 52.24)
};
\addlegendentry{SEL-Wikinews}

\legend{}; 
\end{axis}

\end{tikzpicture}
   \caption{Stratification of results by relative entity position, i.e., whether the first occurrence of a unique mention is within the first $x\%$ of words in the document. \textit{Above:} AP using the RoBERTa-Large pooling model. \textit{Below:} Fraction of positive ground truth labels (i.e., salient entities) in each dataset.}
   \label{fig:analysis_pos}
\end{figure}
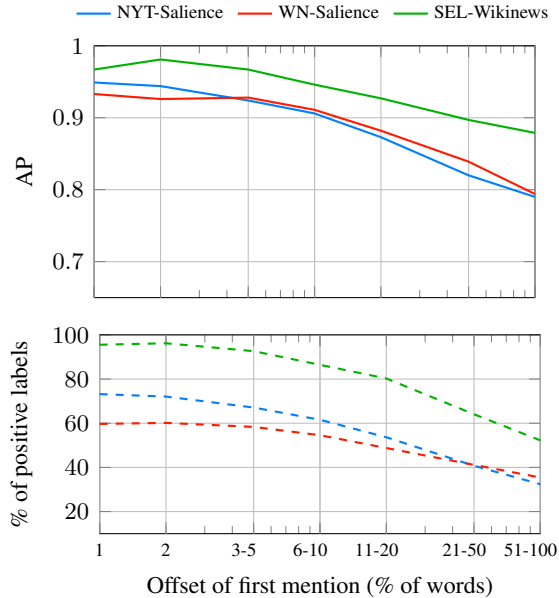

\noindent
\textbf{Position of first entity mention.}
This is a well-studied indicator of entity salience, and is used as a feature in most prior work~\cite{dunietz-gillick-2014-new, trani2018sel, ponza2019swat, wu2020wnsalience}. A positional baseline feature produces the best results on WN-Salience in \citet{wu2020wnsalience}.
Figure~\ref{fig:analysis_pos} illustrates the results stratified by the relative position of (the first token of) the first unique mention of each entity.
We observe that model predictions are most accurate for entities which appear early in the document across all datasets, but AP remains relatively high---around 0.8 or higher---even for entities that appear later.

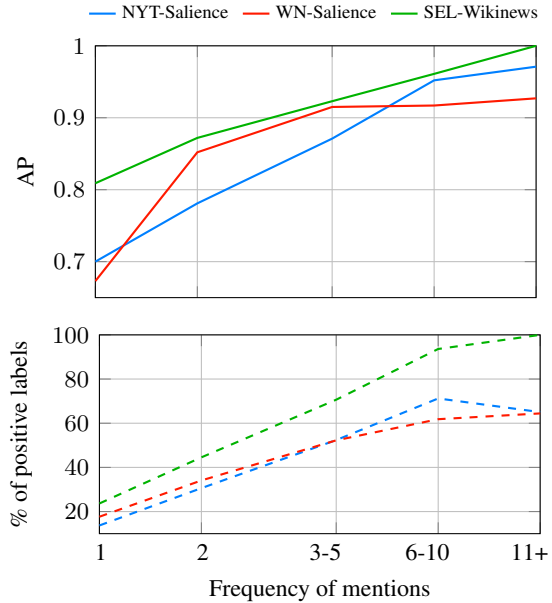
\begin{figure}[t]
\begin{tikzpicture}
    \begin{axis}[
        name=ap,
        legend columns=3,
        legend image code/.code={
            \draw[mark repeat=2,mark phase=2]
            plot coordinates {
            (0cm,0cm)
            (0.2cm,0cm)        
            (0.35cm,0cm)         
            };%
        },
        legend style={at={(0.5,1.05)}, anchor=south, draw=none, font=\scriptsize, /tikz/every even column/.append style={column sep=3pt}},
        small,
        width=210pt,
        height=140pt,
        ylabel={AP},
        xmode=log,
        log ticks with fixed point,
        xtick=data,
        xticklabels={\quad 1, \quad 2, 3-5, 6-10, 11+},
        xticklabel style={font=\footnotesize, xshift=0.0ex, anchor=60},
        xticklabels={\empty},
        xmin=1,
        xmax=20,
        ymin=0.65,
        ymax=1,
        xmajorgrids=true,
        ymajorgrids=true,
]

\addplot [blue!50!cyan, thick] coordinates {
    (1, 0.700)
    (2, 0.781)
    (5, 0.871)
    (10, 0.952)
    (20, 0.971)
};
\addlegendentry{NYT-Salience}

\addplot [red!80!orange, thick] coordinates {
    (1, 0.673)
    (2, 0.852)
    (5, 0.915)
    (10, 0.917)
    (20, 0.927)
};
\addlegendentry{WN-Salience}

\addplot [green!70!black, thick] coordinates {
    (1, 0.809)
    (2, 0.872)
    (5, 0.923)
    (10, 0.961)
    (20, 1.000)
};
\addlegendentry{SEL-Wikinews}
\end{axis}

    \begin{axis}[
        name=posrate,
        at={(ap.outer south)}, anchor=outer north,
        legend columns=3,
        legend style={at={(0.5,1.05)}, anchor=south, draw=none, font=\scriptsize, /tikz/every even column/.append style={column sep=3pt}},
        small,
        width=210pt,
        height=120pt,
        xlabel={Frequency of mentions},
        ylabel={\% of positive labels},
        xmode=log,
        log ticks with fixed point,
        xtick=data,
        xticklabels={\quad 1, \quad 2, 3-5, 6-10, 11+},
        xticklabel style={font=\footnotesize, xshift=0.0ex, anchor=60},
        xmin=1,
        xmax=20,
        ymin=10,
        ymax=100,
        xmajorgrids=true,
        ymajorgrids=true,
]

\addplot [blue!50!cyan, thick, dashed] coordinates {
    (1, 13.71)
    (2, 30.61)
    (5, 52.37)
    (10, 71.21)
    (20, 65.03)
};
\addlegendentry{NYT-Salience}

\addplot [red!80!orange, thick, dashed] coordinates {
    (1, 17.67)
    (2, 34.09)
    (5, 52.25)
    (10, 61.82)
    (20, 64.44)
};
\addlegendentry{WN-Salience}

\addplot [green!70!black, thick, dashed] coordinates {
    (1, 23.67)
    (2, 44.55)
    (5, 70.65)
    (10, 93.61)
    (20, 100.00)
};
\addlegendentry{SEL-Wikinews}

\legend{}; 
\end{axis}

\end{tikzpicture}
   \caption{Stratification of results by the frequency of an entity. \textit{Above:} AP using the RoBERTa-Large pooling model. \textit{Below:} Fraction of positive ground truth labels (i.e., salient entities) in each dataset.}

   \label{fig:analysis_freq}
\end{figure}

\noindent
\textbf{Frequency of mentions.}
This is another intuitive and well-studied indicator of entity salience, used as a feature in most prior work. Figure~\ref{fig:analysis_freq} shows the results stratified by the frequency of each unique entity mention.
We observe, unsurprisingly, that frequently-mentioned entities are easier to classify, but performance remains strong even when classifying very infrequent mentions, e.g., mentions occurring only twice are classified with an AP around or greater than 0.8. Unique mention frequency is an underestimate of \textit{entity} frequency due to coreference errors, and infrequent mentions may consequently be more ambiguous in terms of salience.


\noindent
\textbf{Unseen entities.}
As our models rely on the full text of articles, it is possible that they may overfit on entities seen in training and fail to generalize. Table~\ref{table:unseen} stratifies the results by whether unique mentions occur in the training data. The results show stable performance for NYT-Salience and SEL-Wikinews, but poor performance on unseen entities in WN-Salience. However, we observe that the latter are also \textit{rarely} labeled salient in the ground truth, suggesting that the entity-centric sampling strategy of WN-Salience may miss entities that would be judged as salient by human readers.

\begin{table}[t]
\centering
\small
\setlength{\tabcolsep}{4pt}
\renewcommand{\arraystretch}{1.1}

\begin{tabular}{l|cr|cr|cr} 
\hline
{Entities} & \multicolumn{2}{c|}{NYT-Salience} & \multicolumn{2}{c|}{WN-Salience} & \multicolumn{2}{c}{SEL-Wikinews} \\
& AP & \% +ve & AP & \% +ve  & AP & \% +ve  \\
\hline
\hline
Seen & 0.787 & 32.7\% & \textbf{0.821} & 46.2\% & \textbf{0.900} & 55.9\% \\ 
Unseen & \textbf{0.817} & 29.1\% & 0.188 & 4.4\% & 0.862 & 47.3\% \\ 
\hline 
All & 0.794 & 32.4\% & 0.790 & 35.4\% & 0.879 & 52.2\% \\ 
\hline
\end{tabular} 
\caption{Stratification of results by whether the entity mention was seen in the training data or not. The values reported for each group of entities are AP and the fraction of positive ground truth labels (i.e., salient entities).}
\label{table:unseen}
\end{table}

\section{Conclusion}
 We find that pretrained transformer models can be fine-tuned to yield representations that robustly encode entity salience. We propose a variation on the typical classification heads for these models to estimate the salience of all candidate entities with a single encoding of the document, thereby avoiding the computational overhead of repeated inference passes for each candidate entity.
 Distilling with teacher ensembles produces small student models that perform as well as models with $4\times$ the number of parameters. These models also appear to capture classic signals of salience such as mention position and frequency, and avoid overfitting to entities in the training data. Finally, we demonstrate that temperature scaling during distillation can modulate the calibration of student models, thereby providing a simple solution for the miscalibrated predictions often encountered with deep neural networks. 


\section*{Limitations}
Because two of the benchmark datasets used in this work were constructed using automated rules from the New York Times corpus and Wikinews, the results obtained here may not carry over to tasks involving less rigorous definitions of entity salience. Moreover, the news articles in these datasets (without entity salience labels or derivation rules) may have been included in the pretraining corpus for the RoBERTa models we used, including those from the test partitions. As a mitigating factor for such concerns around generalization, we also conducted experiments on a private dataset of approximately 5,400 news articles published after 2021 containing manual annotations of salience for each entity. These experiments produced similar results and ablation patterns as those presented in this work, but we are currently unable to release the dataset.

The use of a pretrained masked language model for classification also raises questions of bias in prediction, both in the general case where salience labels may be associated with irrelevant attributes of an entity such as ethnicity or gender, and in the specific case where certain entities may be more or less likely to be judged salient owing to their occurrence in training or pretraining corpora.
 We investigated the latter point with the manually-annotated dataset mentioned above and found there was almost no correlation between the accuracy of salience labels for a test entity and its frequency in the training set. However, it remains challenging to quantify the impact of the pretraining corpus on prediction quality and robustness.

\section*{Ethics Statement}
This work describes the development of models for identifying salient entities in news, which can be used in order to characterize news content for recommendation and ranking systems. However, automated judgments of entity salience are a fairly limited form of article representation---unable to capture news events without established names, prone to encode biases from training or pretraining corpora, and liable to degrade in precision over time as news content evolves---and could thus skew the perceptions and preferences of newsreaders if deployed without sufficient human oversight.

While the development of computationally-efficient salience detection models is the main aim of this research, their strong performance can be partially attributed to the use of multi-teacher knowledge distillation which requires significant computational resources at training time. Consequently, organizations without access to similar resources may be unable to create or maintain such models without sacrificing accuracy.


\bibliography{anthology,custom}

\newpage
\newpage
\appendix

\begin{table*}[!tb]
\centering
\small
\renewcommand{\arraystretch}{1.1}
\begin{tabular}{lr|cccc|cccc} 
\hline
\multirow{2}{*}{Model} &  & \multicolumn{4}{c|}{WN-Salience} & \multicolumn{4}{c}{SEL-Wikinews} \\
& & P & R & F$_1$ & AP & P & R & F$_1$ & AP \\
\hline
\hline
\multicolumn{2}{l|}{\citet{wu2020wnsalience}}  & 0.479 & 0.532 & 0.504 & - & 0.403 & \textbf{0.856} & 0.548 & -   \\
\hline
\multirow{3}{*}{DistilRoBERTa} & Standard & 0.670 & 0.720 & 0.694 & 0.765 & 0.692 & 0.680 & 0.686 & 0.733 \\
& Tagging  &  0.599 & 0.710 & 0.650 & 0.732 & 0.737 & 0.812 & 0.773 & 0.841  \\
& Pooling  & 0.664 & 0.701 & 0.682 & 0.760 & 0.763 & 0.802 & 0.782 & 0.841   \\
\hline
\multirow{3}{*}{RoBERTa-Base} & Standard & 0.696 & 0.723 & 0.709 & 0.778 & 0.734 & 0.766 & 0.750 & 0.813 \\
& Tagging & 0.647 & 0.721 & 0.682 & 0.763 & 0.774 & 0.782 & 0.778 & 0.860   \\
 & Pooling & 0.657 & 0.725 & 0.689 & 0.772 & 0.797 & 0.838 & 0.817 & 0.859   \\
\hline
\multirow{3}{*}{RoBERTa-Large} & Standard & 0.688 & \textbf{0.768} & \textbf{0.726} & 0.788 & 0.777 & 0.838 & 0.806 & 0.861 \\
& Tagging & 0.715 & 0.690 & 0.703 & 0.789 & \textbf{0.802} & 0.843 & 0.822 & 0.865  \\
 & Pooling & \textbf{0.720} & 0.705 & 0.712 & \textbf{0.794} & 0.796 & 0.853 & \textbf{0.824} & \textbf{0.876}  \\
\hline
\end{tabular} 
\caption{Performance of the proposed pooling and tagging approaches for RoBERTa models of different sizes, including a standard binary classifier that re-encodes the document for each entity mention.}
\label{table:standard}
\end{table*}

\section{Comparison to Standard Classifiers}
\label{sec:standard}
A key advantage of our proposed models is that they only need to encode the input document once, avoiding the computational overhead of a full forward pass for every entity in a news article. To examine how this affects accuracy, we also conducted experiments with simpler RoBERTa-based binary classifiers which re-encode the document for each candidate. Here, the input consists of an entity mention followed by the text of the article, separated by a special separator token. The contextual representation of the standard classification token \texttt{<CLS>} is then fed to a binary classification head to obtain the salience score for that entity.

We conducted this experiment on the smaller datasets WN-Salience and SEL-Wikinews, as standard classification requires long training schedules to converge on NYT-Salience. The results in Table~\ref{table:standard} show that the proposed pooling models are competitive with standard RoBERTa classification on WN-Salience, with $<1\%$ difference in AP. On SEL-Wikinews, pooling outperforms the standard approach by around $11\%$ for DistilRoBERTa, $4\%$ for RoBERTa-Base and $2\%$ for RoBERTa-Large, suggesting that fine-tuning with pooling may be advantageous when training datasets are small.

These results also show that standard fine-tuning is not worth the additional computational cost and runtime overhead over our proposed pooling technique. As classification heads comprise a small part of the model and require negligible runtime compared to the encoding of the input document, the runtime speedup for pooling over standard classification scales in proportion to the number of entities in a document. As estimated in Table~\ref{table:speedup}, pooling is an order of magnitude faster than standard binary classification for all three datasets.


\begin{table}[t]
\centering
\small
\renewcommand{\arraystretch}{1.1}
\begin{tabular}{lc}
\hline
\multirow{2}{*}{Dataset} & Runtime speedup of \\
 & Pooling vs Standard \\
\hline
NYT-Salience & $20.0\times$ \\
WN-Salience & $12.4\times$ \\
SEL-Wikinews & $27.9\times$ \\
\hline
\end{tabular}
\caption{Approximate decrease in runtime of pooling relative to standard classification on average for each dataset, inferred from entity statistics in Table \ref{table:corpus_stats}.}
\label{table:speedup}
\end{table}

\section{Prompting LLMs for Entity Salience}
\label{sec:llms}
This paper describes techniques for fine-tuning pretrained transformer encoders---typically trained as masked language models---to produce effective and scalable classifiers of entity salience. However, in recent years, large language models (LLMs) based on autoregressive transformer decoders have demonstrated significant capabilities in few-shot and zero-shot tasks~\cite{brown2020gpt3,openai2023gpt4}. LLMs can be prompted to create new baselines for salience estimation, but we find they have limitations that prevent robust experimentation in our salience classification setting.

First, entity salience is not consistently defined in the datasets we study and it is fairly difficult to characterize precisely, as illustrated by the variance in salience dataset statistics from Table~\ref{table:corpus_stats} as well as the limits to transfer learning across datasets in Appendix~\ref{sec:transferability}. Consequently, it is challenging to construct zero-shot prompting strategies for entity salience that can work effectively across all datasets.
Few-shot prompting can make use of a small number of labeled examples to guide the prediction, but
it does not typically perform as well as fine-tuning~\cite{mosbach-etal-2023-shot,zhang-etal-2023-machine}. For text classification tasks in particular, fine-tuning of smaller BERT-like classifiers can outperform in-context learning with LLMs even in few-shot settings~\cite{mosbach-etal-2023-shot,edwards2024language}, and fine-tuning can naturally better exploit the larger training partitions available for the benchmark datasets used in this study. Few-shot prompting also requires orders-of-magnitude more encoding latency to encode multiple labeled articles for each inference task. 
Furthermore, prior research on in-context learning~\cite{lu-etal-2022-fantastically,min-etal-2022-rethinking}
has shown that classification performance can be highly sensitive to the choice, order and number of few-shot examples provided, thereby implying that a single prompting approach may not be representative of LLM performance on salient entity detection tasks. 

LLMs are also hindered by practical considerations.
For example, if salience predictions are needed for specific candidate entities---such as ones disambiguated through entity linking---an LLM needs to label entities \textit{within} the text rather than just enumerate them as errors may be introduced when mapping results back to candidates. In addition, meaningful scores for salience cannot be easily computed over token predictions unless LLMs are used as binary classifiers, with the contingent scaling challenges noted in Appendix~\ref{sec:standard}.

We therefore conclude that LLMs do not make for reliable baselines in this setting and exclude them from our experiments.
However, the disadvantages noted above can be minimized when using LLMs as teacher models during knowledge distillation, which we intend to explore in future work.
\section{Entity Salience in the Ranking Setting}
\label{sec:ranking}
Although most prior work evaluates entity salience estimation in the classification setting, a branch of research in supervised~\cite{xiong2018towards} and unsupervised~\cite{lu2021salience} salience estimation adopts a framework in which entities are \textit{ranked} by their salience, resulting in model evaluations on the NYT corpus that use top-$k$ ranking measures such as P@$k$, R@$k$ and F$_1$@$k$. However, these works use salience annotations on a much larger portion of the New York Times corpus~\cite{sandhaus08} than the subset of articles annotated by \citet{dunietz-gillick-2014-new} which is used in the classification literature. This larger dataset has $5.9\times$ more training and validation examples (526,126 + 64,000) and $6.5\times$ more test examples (63,589) than the NYT-Salience dataset we use. Consequently our NYT-Salience models cannot be directly compared to models from the literature that were evaluated on NYT data in the ranking setting.

\begin{table}[t]
\centering
\small
\setlength{\tabcolsep}{3.6pt}
\renewcommand{\arraystretch}{1.1}

\begin{tabular}{l|cc|cc}
\hline
Model & P@1 & P@5 & R@1 & R@5 \\
\hline
\hline
\citet{dunietz-gillick-2014-new}$^\dagger$ & 0.639 & 0.461 & 0.089 & 0.285 \\
\citet{xiong2018towards}$^\dagger$ & 0.687 & 0.508 & 0.101 & 0.334 \\
\hline
DistilRoBERTa & 0.849 & 0.463 & 0.421 & 0.892 \\
RoBERTa-Base  & 0.863 & 0.469 & 0.428 & 0.902 \\
RoBERTa-Large & 0.880 & 0.472 & 0.438 & 0.907 \\
\hline
\end{tabular}
\caption{Performance of our pooling models on NYT-Salience under top-$k$ precision \& recall. $\dagger$ indicates results from \citet{xiong2018towards} in which models were trained and evaluated on a much larger NYT dataset and are thus not directly comparable to our models.}
\label{table:kesm}
\end{table}

Table~\ref{table:kesm} presents the performance of our approach on NYT-Salience under top-$k$ precision and recall metrics, as reported in \citet{xiong2018towards}. For context, we also report the performance of key models in their evaluation, but we cannot draw conclusions on relative performance as those models use the much larger NYT dataset noted above.
%
%
%

\begin{table}[t]
\centering
\small
\setlength{\tabcolsep}{3pt}
\renewcommand{\arraystretch}{1.1}

\begin{tabular}{l|cccc} 
\hline
{Model} &  {P} & {R} & {F$_1$} & {AP} \\
\hline
\hline
 \citet{wu2020wnsalience} & 0.479 & 0.532 & 0.504 & -    \\
\hline
 DistilRoBERTa (NYT) & 0.460 & 0.587 & 0.516 & 0.488 \\
RoBERTa-Base (NYT) & 0.450 & 0.620 & 0.521 & 0.491 \\
RoBERTa-Large (NYT) & 0.477 & 0.571 & 0.520 & 0.483 \\
 \hline
DistilRoBERTa (WN) & 0.664 & 0.701 & 0.682 & 0.760    \\
RoBERTa-Base (WN) & 0.657 & 0.725 & 0.689 & 0.772     \\
RoBERTa-Large (WN) & \textbf{0.720} & 0.705 & \textbf{0.712} & \textbf{0.794}  \\
\hline
RoBERTa-Large (NYT+WN) & 0.656 & \textbf{0.733} & 0.693 & 0.768 \\
\hline 
\end{tabular} 
\caption{Performance of pooling models trained on NYT-Salience and evaluated on WN-Salience. The training dataset is noted in parentheses.}
\label{table:transferability}
\end{table}

\section{Transferability across Datasets}
\label{sec:transferability}
To study whether salience prediction can be transferred across datasets, we trained models on NYT-Salience and evaluated them on WN-Salience. The metrics tend to degrade significantly owing to the difference between the notions of salience in each dataset, but the results are still better than the baseline model~\cite{wu2020wnsalience} as indicated in Table~\ref{table:transferability}. Training a RoBERTa-Large pooling model on combined training data performs comparably well on both NYT-Salience and WN-Salience, suggesting the models ability to adapt to different feature and label distributions. From a separate experiment, we also observed that training DistilRoBERTa models on NYT-Salience and fine-tuning/re-training on WN-Salience leads to slight improvements over just fine-tuning the pretrained DistilRoBERTa model.

\section{Improving Student Model Calibration}
\label{sec:calibration}
An important use case of salience scores is to provide input features to downstream systems for content understanding and personalization. These systems often rely on the raw output score of entity salience models, making them sensitive to the score distribution of the model. To ensure that these scores are interpretable and to keep them as consistent as possible across model versions,
the best approach is to ensure that the models scores are \emph{calibrated} \cite{guo2017calibration}. A calibrated model is one for which its output confidence score is equal to the empirical probability of the predicted category, e.g., if a model predicts a score of 0.7, then there is a 70\% chance that the predicted category is correct.
Here, we measure how calibration varies as we adjust the temperature scaling hyperparameters during knowledge distillation and show that calibration can be preserved without post-processing, albeit with some loss in accuracy. 

\subsection{Measuring Calibration}
The calibration of salience models is measured using the expected calibration error (ECE) metric \cite{naeini2015obtaining}. 
ECE is the weighted average of absolute difference between models accuracy and confidence, using a histogram of the model score distribution. 
\begin{align}
\label{eq:ece}
    \text{acc}(B_m) &= \frac{1}{|B_m|} \sum_{i\in B_m} \mathbbm{1}(\hat{y}_i=y_i) \\
    \text{conf}(B_m) &= \frac{1}{|B_m|} \sum_{i\in B_m} p_i \\
    \text{ECE} &= {\sum_{m=1}^{M} \frac{|B_m|}{n}} |\text{acc}(B_m) - \text{conf}(B_m)|
\end{align}
where $|B_m|$ is the number of samples in the $m$th of $M$ histogram bins, and $n$ is the total number of samples available. Here, $\text{acc}(B_m)$ denotes the accuracy of predictions in bin $B_m$ and $\text{conf}(B_m)$ denotes the average of predictions in $B_m$.

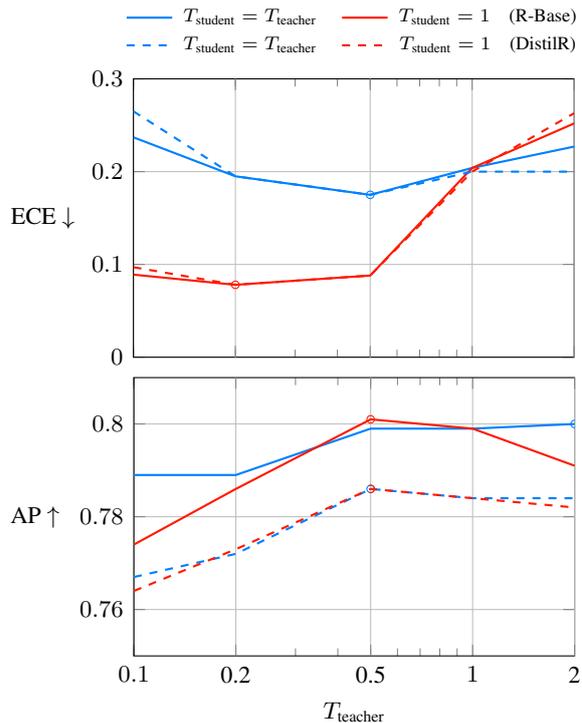
\begin{figure}[t]
\begin{tikzpicture}
    \begin{axis}[
        name=ece,
        legend columns=2,
        legend style={at={(0.5,1.05)}, anchor=south, draw=none, font=\scriptsize, /tikz/every even column/.append style={column sep=7pt}},
        small,
        width=210pt,
        height=150pt,
        ylabel={ECE $\downarrow$},
        ylabel style={rotate=-90, at={(axis description cs:-0.11,0.5)}},
        xmode=log,
        log ticks with fixed point,
        extra x ticks={0.2, 0.5, 2},
        extra x tick labels={},
        xticklabels={\empty},
        xmin=0.1,
        xmax=2.0,
        ymin=0.0,
        ymax=0.3,
        xmajorgrids=true,
        ymajorgrids=true,
]

\addplot [blue!50!cyan, thick] coordinates {
    (0.1, 0.237)
    (0.2, 0.195)
    (0.5, 0.175)
    (1.0, 0.204)
    (2.0, 0.227)
};
\addlegendentry{\ $T_\text{student} = T_\text{teacher}$}

\addplot [red!80!orange, thick] coordinates {
    (0.1, 0.089)
    (0.2, 0.078)
    (0.5, 0.088)
    (1.0, 0.204)
    (2.0, 0.252)
};
\addlegendentry{\ $T_\text{student} = 1$\quad (R-Base)}

\addplot [blue!50!cyan, thick, dashed] coordinates {
    (0.1, 0.265)
    (0.2, 0.195)
    (0.5, 0.175)
    (1.0, 0.200)
    (2.0, 0.200)
};
\addlegendentry{\ $T_\text{student} = T_\text{teacher}$}

\addplot [red!80!orange, thick, dashed] coordinates {
    (0.1, 0.097)
    (0.2, 0.078)
    (0.5, 0.088)
    (1.0, 0.200)
    (2.0, 0.263)
};
\addlegendentry{\ $T_\text{student} = 1$\quad (DistilR)}

\node [draw, blue!50!cyan, circle, inner sep=1pt] 
      at (axis cs: 0.5, 0.175) {};
\node [draw, red!80!orange, circle, inner sep=1pt] 
      at (axis cs: 0.2, 0.078) {};
\end{axis}

    \begin{axis}[
        name=ap,
        at={(ece.outer south)}, anchor=outer north,
        legend columns=2,
        legend style={at={(0.5,1.0)}, anchor=south, draw=none, font=\scriptsize, /tikz/every even column/.append style={column sep=10pt}},
        small,
        width=210pt,
        height=150pt,
        xlabel={$T_\text{teacher}$},
        ylabel={AP $\uparrow$},
        ylabel style={rotate=-90},
        xmode=log,
        log ticks with fixed point,
        extra x ticks={0.2, 0.5, 2},
        extra x tick labels={0.2, 0.5, 2},
        xmin=0.1,
        xmax=2.0,
        ymin=0.75,
        ymax=0.81,
        xmajorgrids=true,
        ymajorgrids=true,
]

\addplot [blue!50!cyan, thick] coordinates {
    (0.1, 0.789)
    (0.2, 0.789)
    (0.5, 0.799)
    (1.0, 0.799)
    (2.0, 0.800)
};
\addlegendentry{\ $T_\text{student} = T_\text{teacher}$}

\addplot [red!80!orange, thick] coordinates {
    (0.1, 0.774)
    (0.2, 0.786)
    (0.5, 0.801)
    (1.0, 0.799)
    (2.0, 0.791)
};
\addlegendentry{\ $T_\text{student} = 1$\quad (R-Base)}

\addplot [blue!50!cyan, thick, dashed] coordinates {
    (0.1, 0.767)
    (0.2, 0.772)
    (0.5, 0.786)
    (1.0, 0.784)
    (2.0, 0.784)
};
\addlegendentry{\ $T_\text{student} = T_\text{teacher}$}

\addplot [red!80!orange, thick, dashed] coordinates {
    (0.1, 0.764)
    (0.2, 0.773)
    (0.5, 0.786)
    (1.0, 0.784)
    (2.0, 0.782)
};
\addlegendentry{\ $T_\text{student} = 1$\quad (DistilR)}

\node [draw, blue!50!cyan, circle, inner sep=1pt] 
      at (axis cs: 2.0, 0.800) {};
\node [draw, red!80!orange, circle, inner sep=1pt] 
      at (axis cs: 0.5, 0.801) {};
\node [draw, blue!50!cyan, circle, inner sep=1pt] 
      at (axis cs: 0.5, 0.786) {};
\node [draw, dashed, red!80!orange, circle, inner sep=1pt] 
      at (axis cs: 0.5, 0.786) {};

\legend{}; 
\end{axis}

\end{tikzpicture}
   \caption{Calibration (ECE $\downarrow$) and average precision (AP $\uparrow$) metrics of distilled pooling models as the teacher model temperature $T_\text{teacher}$ is varied during distillation on the WN-Salience dataset. Blue lines show metrics when the student model temperature $T_\text{student}$ is set equal to $T_\text{teacher}$, while red lines show $T_\text{student}$ fixed to 1. Solid lines show RoBERTa-Base student models while dashed lines indicate DistilRoBERTa models. Highlighted points ($\circ$) indicate the best result for each plot.}
   \label{fig:ece_ap}
\end{figure}

\subsection{Calibration via Temperature Scaling}
\citet{hinton2015distilling} formulate knowledge distillation by applying a high temperature on the logits of student and teacher models during training. In this work, we experiment with different temperature scaling combinations for both student and teacher. Figure~\ref{fig:ece_ap} summarizes the results for the effect that temperature scaling during knowledge distillation has on the performance and calibration of distilled pooling models on the WN-Salience dataset. We find that the conventional approach of using higher temperatures (around 1 or greater) for both teacher and student generally result in improved student model performance in comparison to using lower temperatures or hard labels, but this does not lead to well-calibrated models.

To improve calibration, we reduced the temperature of \textit{only} the teacher model logits to bring them closer to hard labels, while keeping student model temperatures fixed to 1. This leads to more calibrated and more accurate models than those trained on hard labels alone. This setting illustrates the tradeoff between model performance and calibration: more accurate models are often less calibrated, but the tradeoff can be controlled by tuning the teacher temperature. Model scores can be further calibrated if necessary using methods such as isotonic regression or by applying an additional temperature scaling fine-tuning step \cite{guo2017calibration}.

\end{document}